\documentclass[journal]{IEEEtran}
\usepackage{mathrsfs}
\usepackage{graphicx, graphics, times, amsfonts, amsmath, amssymb, cite}
\usepackage{amsthm}
\usepackage{mathtools}
\usepackage{color}
\usepackage{hyperref}
\usepackage{multirow}
\usepackage{rotating}
\usepackage[font=small]{caption}
\usepackage{subcaption}
\usepackage{bm}

\input{mysymbol.sty}
\usepackage[utf8]{inputenc}
\usepackage{enumitem}
\usepackage[ruled,linesnumbered]{algorithm2e}
\usepackage[capitalise]{cleveref}
\usepackage{comment}

\usepackage{tikz, pgfplots}
\pgfplotsset{compat=1.17}
\pgfplotstableset{col sep=semicolon}
\usepackage{moresize} 
\usepgfplotslibrary{groupplots}
\usepgfplotslibrary{external}
\tikzset{every mark/.append style={scale=1.4, solid}, font=\footnotesize}
\pgfplotsset{
    width=1\textwidth,
    height=5.5cm,
    legend style={
        font=\ssmall ,  
        inner xsep=1pt,
        inner ysep=1pt,
        nodes={inner sep=1pt}},
    legend cell align=left,
    every axis/.append style={line width=0.5pt},
 	every axis plot/.append style={line width=1.25pt},
 	every axis y label/.append style={yshift=-4pt}
}

\newcommand{\Tr}{\mathsf{T}}

\begin{document}

\title{Exploiting the Structure of Two Graphs with Graph Neural Networks}

\author{Victor M. Tenorio,~\IEEEmembership{Student Member,~IEEE},
        and~Antonio G. Marques,~\IEEEmembership{Senior Member,~IEEE} \vspace{-0.4cm} 
\thanks{V. M. Tenorio and A. G. Marques are with the Department
of Signal Theory and Comms., King Juan Carlos University, Madrid, Spain, \{victor.tenorio,antonio.garcia.marques\}@urjc.es.}
\thanks{This work was supported in part by the Spanish AEI (10.13039/501100011033) under Grants PID2022-136887NBI00, TED2021-130347B-I00 and FPU20/05554, the Young Researchers R\&D Project under ref. num. F861 (CAM and URJC), and the Community of Madrid within the ELLIS Unit Madrid framework.}}


\maketitle

\begin{abstract}
As the volume and complexity of current datasets continue to increase, there is an urgent need to develop new deep learning architectures that can handle such data efficiently.
Graph neural networks (GNNs) have emerged as a promising solution to deal with unstructured data, outperforming traditional deep learning architectures. However, most of the current GNN models are designed to work with a single graph, which limits their applicability in many real-world scenarios where multiple graphs may be involved.
To address this limitation, we propose a novel graph-based deep learning architecture to handle tasks where two sets of signals exist, each defined on a different graph. We first address the supervised and semi-supervised cases, where the input is represented as a signal on top of one graph (the input graph) and the output is a graph signal defined over a different graph (the output graph). For this setup, we propose a three-block architecture where we first process the input data using a GNN that operates over the input graph, then apply a transformation function that operates in a latent space and maps the signals from the input to the output graph, and finally implement a second GNN that operates over the output graph. Our goal is not to propose a single specific definition for each of the three blocks, but rather to provide a flexible approach to solve tasks involving data defined on two graphs. The second part of the paper addresses a self-supervised setup, where the focus is not on the output space but on the underlying latent space, and, inspired by Canonical Correlation Analysis, we seek informative representations of the data that can be leveraged to solve a downstream task. By leveraging information from multiple graphs, the proposed architecture can capture more intricate relationships between different entities in the data, leading to enhanced learning performance in a wide range of applications. We test this in several experimental setups using synthetic and real world datasets, and observe that the proposed architecture works better than traditional deep learning architectures, showcasing the importance of leveraging the information of the two graphs involved in the task.
\end{abstract}

\section{Introduction}\label{S:Introduction}

\IEEEPARstart{S}{cientific} and social progress requires processing, understanding, and learning from vast amounts of heterogeneous signals, demanding new models and tools to handle the irregular structure and nonlinear interactions of contemporary data.
A popular and versatile approach consists in modeling the underlying irregular structure of the signals as a graph, a strategy with applications ranging from neurological activity defined on brain networks to social networks where epidemics spread~\cite{shuman2012theemergingfield,kolaczyk2009book}. 
This is precisely the approach put forth in Graph Signal Processing (GSP)~\cite{djuric2018bookGSP,marques2020editorial}, a field devoted to generalizing classical tools from signal processing to signals defined over graphs.
Early efforts investigated linear graph filters, graph Fourier representations, and inverse problems involving graph signals~\cite{SandryMouraSPG_TSP14Freq}.
More recently, the trend has shifted interest toward higher-dimensional graph signals and learning non-linear relations between the signals and the graph.
As a result, the development of graph-aware architectures, usually known as Graph Neural Networks (GNNs)~\cite{bronstein2017geometricdeeplearning,tenorio2021robust}, has attracted significant attention well beyond the machine learning and signal processing communities, with deep convolutional architectures being placed among the most prominent alternatives~\cite{defferrard2016convolutional,kipf2017semisupervised}.

While GNNs have proven to be powerful tools for handling signals whose underlying structure may be represented as a single graph, many real-world applications require multiple graphs to accurately represent the complex relationships within the data.
Traditionally, GNNs use graph signals representing vectors of node features as their input, and the output is either a single unidimensional label or a signal defined over the same graph. For the latter case, which is the one we address in this paper, the output can represent either a signal of interest (in a regression task) or a vector of node labels (in a node classification setting).
However, multiple graphs may be needed, for example, in recommendation systems, where the user-user and item-item interactions are better represented using two different graphs.
Another scenario where multiple graphs naturally appear arises in the field of Computational Fluid Dynamics (CFD), where simulations over a fine mesh can be computationally costly.
To reduce the computational burden, one can perform the simulations over a coarse mesh and then interpolate the values to a finer mesh~\cite{belbute20solvers}.
Deep learning approaches have shown potential to address the interpolation task with moderate computational complexity, so resorting to GNNs accounting for both the topology of the coarse and fine meshes appears as a promising alternative.

In a different but related context, Canonical Correlation Analysis (CCA)~\cite{Hardoon2004CanonicalCA} aims to project two views of the same data (and of possibly different dimensionality) to a low-dimensional common latent space. These projections aim to maximize the correlation between the two transformed views while adhering to orthonormality constraints.
Previous works have extended CCA to graphs~\cite{chen18graphcca,chen19gmcca,kaloga21VGAE}, but they consider the two views as different signals defined over the same graph.
However, in real world scenarios, we may encounter situations where the two views are defined over different graphs, rendering the current state-of-the-art architectures impossible to use.
Indeed, classical CCA assumes that the covariances of the views are different, hence the natural generalization to the graph domain is to work with different graphs for each of the views.
The transformed views obtained by CCA have been used in the past in the context of, e.g., Self-Supervised Learning (SSL)~\cite{xie23sslreview,liu22sslsurvey}, where we learn alternative rich representations of our data without using labels, and then use these representations and a small number of labeled samples to train a simpler classifier like a logistic regressor. This methodology, particularly pertinent when the labeled set is small, may lead to a performance comparable to that of a supervised setting~\cite{zhang2021from}.


To address the aforementioned limitations, we propose a graph-based architecture tailored to tasks where we have pairs of graph signals $\{\bbX_p,\bbY_p\}_{p=1}^P$, with $\bbX_p$ being a signal defined over the (input) graph $\ccalG_X$ with $N_X$ nodes, and $\bbY_p$ being a signal defined over the (output) graph $\ccalG_Y$ with a possibly different number of nodes $N_Y$.
In the supervised case, the goal is to learn a mapping from $\bbX_p$ to $\bbY_p$ using a training set.
To achieve this, we propose a three-block architecture where the first block implements a GNN defined over the input graph $\ccalG_X$. The output of this GNN is still defined over the input graph, so the second block seeks to transform signals of dimension $N_X$ to signals of dimension $N_Y$, which can be associated with the nodes of the output graph. This transformation can be fixed in advance or learned from the data. Finally, the third block of our architecture applies another GNN over the output graph $\ccalG_Y$ to generate the output graph signal. In contrast to current alternatives, the proposed GNN leverages the information contained in both graphs to learn the mapping from the input signals to the output signals. Furthermore, we also use the proposed architecture to address the unsupervised case where the goal is to learn a transformation that (jointly) maps $\bbX_p$ and $\bbY_p$ to a common latent space. 

More precisely, the contributions of the present work are the following:
\begin{itemize}
    \item We introduce a novel framework to deal with tasks whose input are signals on top of a graph, but whose output is defined over a different graph. This is achieved by a combination of two different GNNs and a transformation function that maps the features associated with nodes of the input graph to features associated with nodes of the output graph. The architecture is trained end-to-end and, during the operation phase, it is used to estimate the output graph signal for a given input graph signal.
    \item We propose different alternatives for the GNNs and, critically, for the transformation function. These are both application agnostic, which can be applied in any setting, or domain specific, which can be tailored to a specific application.
    \item In the context of unsupervised learning and SSL, we analyze a slightly different setting where the focus is not on the output but on the intermediate latent space, implementing a CCA-based loss function. Once again, the architecture is trained end-to-end and, during the operation phase, the architecture yields a compressed version of the two graph signals (views) used as input (each defined on a different graph). 
    \item We provide a diverse set of numerical experiments with different applications where instances of our architecture offer better performance than alternatives not using the information in both the input and output graphs.
\end{itemize}

The remainder of the paper is organized as follows. We finalize this introductory section by reviewing previous works dealing with multiple graphs. In Section~\ref{S:fundamentals} we introduce the necessary concepts from GSP, GNNs and SSL. In Section \ref{S:methodology} we introduce the proposed framework, while in Section \ref{S:cca_ssl} we explore its connections with the field of CCA and SSL. Section~\ref{S:experiments} presents the numerical experiments developed to highlight different applications of the proposed framework. Finally, Section~\ref{S:conclusion} concludes the paper.

\subsection{Related works}

Working with more than one graph is common in the literature. However, most prior works only change the active edges, while keeping the node set fixed. This is oftentimes referred to as working with multilayer graphs, where each layer comprises a different edge set. GNN architectures tailored for such scenarios have been proposed, such as Tensor GNNs~\cite{ioannidis20tensor} or Multilayer GNNs~\cite{grassia2021mgnn}. Additionally, related bodies of work explore the transferability of GNN architectures across different graphs~\cite{ruiz20graphon,ruiz23transferability}.

Some works focus specifically on considering only two different graphs. For instance, in the protein interface prediction problem, the authors of~\cite{fout17protein} utilize two distinct graphs—one for the ligand protein and another for the receptor protein. Their objective is to classify pairs of nodes from these separate graphs. In another context, \cite{Li2021LearningTM} processes brain signals using two different graphs with the same node set (representing regions of the brain) but distinct edge sets: structural (axonal bundles) and functional (pairwise statistical correlation between the activity of the regions) connections.

Regarding the field of SSL on graphs, several existing works have explored the possibility of learning representations without the use of labels~\cite{xie23sslreview,liu22sslsurvey}. Within this field, some existing works employ a technique of creating two copies of the same graph through random augmentation to facilitate the learning of these representations. Some of these works lie in the field of contrastive graph learning~\cite{hassani20mvrlg,zhu20grace,zhu21gca}, and some of them use a CCA inspired loss function~\cite{zhang2021from}. In both cases, the augmentation keeps the nodal set fixed, while modifying the edge set (via edge dropping) and the node features (via feature masking).

This previous review demonstrates that, while there exists a growing interest in handling scenarios with multiple graphs, existing works focus on graphs with the same nodal set, and current research struggles with leveraging the information in different graphs with different nodes. Notably, our previous conference papers~\cite{rey19encdec,rey20encdec} are two of the few works in the literature addressing the mapping of signals defined over different graphs. These papers propose a GNN-based architecture that relates the input and the output graph using graph pooling. In contrast, this work presents a more comprehensive framework that enables the mapping of graph signals between a pair of graphs without resorting to pooling (which is computationally costly and not straightforward in the graph setting) and employing conventional GNN architectures, which are well-known for their superior performance in various tasks. We also explore the connections between our architecture, CCA and SSL over graphs, and provide real-world numerical experiments that showcase the effectiveness of the proposed framework.

\section{Notation and Fundamentals} \label{S:fundamentals}

In this section we introduce basic principles about graphs, GSP, GNNs, CCA, and its NN-based counterpart, Deep CCA (DCCA).

\vspace{0.2cm}
\noindent
\textbf{GSP Fundamentals}:
Let $\ccalG = (\ccalV, \ccalE)$ represent a graph, where the set $\ccalV$ corresponds to its $|\ccalV| = N$ nodes, and the set $\ccalE$ represents its edges connecting pairs of nodes.
The structure of the graph is represented by its adjacency matrix $\bbA \in \reals^{N \times N}$, whose $(n,n')$ entry $A_{nn'}$ contains the weight associated with the edge $(n',n) \in \ccalE$ and is 0 if $(n',n) \not\in \ccalE$.
The Graph Shift Operator (GSO) $\bbS$ is also a matrix representing the local structure of the graph, whose $(n,n')$-th entry $S_{nn'}$ can be non-zero if $n=n'$ or if $(n',n) \in \ccalE$, without making any assumption about the values in the entries. Some examples of GSO matrices are the adjacency matrix or the combinatorial graph Laplacian $\bbL$~\cite{SandryMouraSPG_TSP14Freq}.
As the GSO encapsulates the graph structure, its application to a graph signal $\bby = \bbS \bbx$ leads to a weighted linear combination of the values of $\bbx$ in the 1-hop neighborhood of each node, and therefore the values of $\bby$ can be computed locally. 
Finally, a graph filter~\cite{segarra2017optimal} is a graph-aware linear operator that is usually represented as a polynomial of the GSO $\bbH = \sum_{r=0}^{R-1} h_r \bbS^r$, with $\bbh = [h_0, \ldots, h_{R-1}] \in \reals^R$ representing the filter coefficients.

\begin{figure*}[t]
    \centering
    \includegraphics[width=\textwidth]{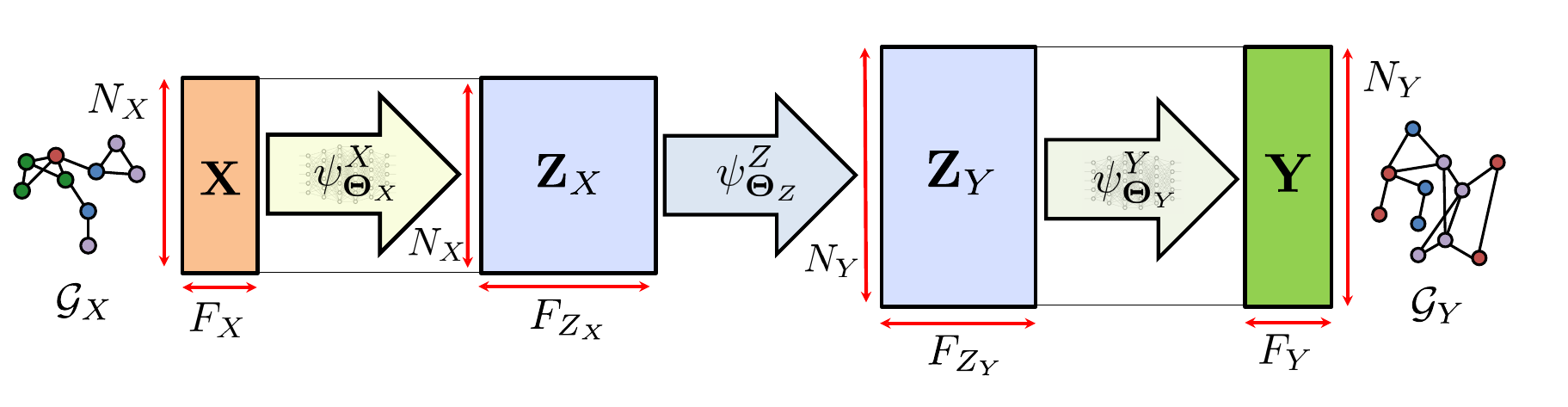}
    \caption{High-level overview of the proposed architecture. The input signal, defined over the input graph $\ccalG_X$, gets processed by the GNN architecture $\psi^X_{\bbTheta_X}$. After that, the underlying space, still defined over $\ccalG_X$ is transformed by means of the function $\psi^Z_{\bbTheta_Z}$ so that it is defined over the output graph $\ccalG_Y$. After that, a new GNN architecture $\psi^Y_{\bbTheta_Y}$ is applied using $\ccalG_Y$ to get the output signal.}
    \label{fig:arch}
\end{figure*}

\vspace{0.2cm}
\noindent
\textbf{GNNs}:
Neural Networks (NNs) are nonlinear architectures composed of layers, where each layer performs a learnable linear combination followed by a pointwise nonlinear function. NNs have achieved remarkable success in the latest years due to their great performance in data based tasks and the simplicity of the training process by means of Gradient Descent (GD) and backpropagation.

GNNs are a specific case of NNs where the linear transformation takes the graph into account. In their most simple form, they aggregate the information in the neighborhood of each node and then combine this information through a learnable transformation. The general case of a GNN is represented by a bank of graph filters~\cite{ruiz2021graph} and is implemented by the recursion
\begin{equation}
    \bbX_{\ell} = \sigma \left( \sum_{r=0}^{R-1} \bbS^r \bbX_{\ell - 1} \bbTheta_{\ell,r} \right),
    \label{eq:gnn_gf}
\end{equation}
where $\bbX_\ell \in \reals^{N \times F_\ell}$ are the graph signals (defined over the graph $\ccalG$) representing the embeddings of layer $\ell$, $\bbS$ represents the GSO of $\ccalG$, $\sigma$ represents a nonlinear activation function applied pointwise and $\bbTheta_{\ell,r}$ represent the learnable parameters of the network, i.e. the coefficients of the filters in the bank.
The most commonly used GNN architecture, referred to by its authors as a Graph Convolutional Network (GCN)~\cite{kipf2017semisupervised}, is a special case of~\eqref{eq:gnn_gf} with filters of order 1, which act as a nonlearnable low-pass graph filter and is implemented by the following recursion
\begin{equation}
    \bbX_{\ell} = \sigma \left( \hbA \bbX_{\ell - 1} \bbTheta_\ell \right),
    \label{eq:gcnn}
\end{equation}
where $\hbA$ is the normalized diagonally loaded adjacency matrix computed as $\hbA = \hbD^{-1/2} (\bbA + \bbI) \hbD^{-1/2}$, with $\hbD = \text{diag} ( (\bbA + \bbI) \bbone)$.

Clearly, in~\eqref{eq:gnn_gf} and~\eqref{eq:gcnn} both the input $\bbX_{\ell-1}$ and the output $\bbX_\ell$ of the layers consist of multi-feature graph signals defined over the same graph $\ccalG$. This is the case for many other GNN architectures~\cite{tenorio2021robust,gama2020graphs,rey2024redesigning}. As a representative example that will be used in the experiment, Graph Attention Networks (GATs)~\cite{velickovic2018graph}, which consider attention mechanisms and give different weights to different edges, also define their input and output as signals defined over the same graph. In general, addressing the consideration of different graphs in a GNN, or incorporating a second graph, is nontrivial and it is the focus of our paper.

\vspace{0.2cm}
\noindent
\textbf{CCA, DCCA and SSL}: 
CCA is a workhorse tool for linear dimensionality reduction and statistical inference when dealing with multiple related datasets, each of them defined on a different domain~\cite{hotelling36cca}.
CCA starts with two views of the same data of possibly different dimension ($\{\bbx_p \in \reals^{N_X}\}_{p=1}^P$ and $\{\bby_p \in \reals^{N_Y}\}_{p=1}^P$), and aims to linearly transform them to a common latent space via the matrices $\bbU \in \reals^{N_Z \times N_X}$ and $\bbV \in \reals^{N_Z \times N_Y}$. In this shared space, the transformed views $\bbU\bbx_p$ and $\bbV\bby_p$ aim to achieve maximal cross-correlation while being orthonormal. Formally, CCA addresses this objective through the optimization problem
\begin{align}
    \label{eq:cca}
    \max_{\bbU, \bbV} \; & \text{tr} ( \bbU \bbSigma_{XY} \bbV^\Tr ) \\
    \text{s. to:} \; & \bbU^\Tr \bbSigma_{XX} \bbU = \bbI \nonumber \\
    & \bbV^\Tr \bbSigma_{YY} \bbV = \bbI, \nonumber
\end{align}
where $\text{tr}$ represents the trace operator, $\bbSigma_{XY} := \frac{1}{P}\sum_{p=1}^P \bbx_p \bby_p^\Tr \in \reals^{N_X \times N_Y}$ is the sample cross-covariance matrix and $\bbSigma_{XX} := \frac{1}{P} \sum_{p=1}^P\bbx_p \bbx_p^\Tr \in \reals^{N_X \times N_X}$ (respectively $\bbSigma_{YY}$) is the covariance matrix of the first (second) view. This problem has a closed-form solution, given by the Singular Value Decomposition (SVD) of the  matrix $\bbT = \bbSigma_{XX}^{-1/2} \bbSigma_{XY} \bbSigma_{YY}^{-1/2}$ \cite{Hardoon2004CanonicalCA}.

Due to the considerable success of NNs, alternative deep approaches to CCA, where the linear transformation is replaced with an NN, have emerged~\cite{andrew13dcca}. However, the straightforward closed-form solution of linear CCA is not achievable in this setting due to the inherent nonlinear nature of NN architectures. The authors of~\cite{andrew13dcca} formulate the DCCA problem as
\begin{align}
    \max_{\bbTheta_X, \bbTheta_Y} \; & \frac{1}{P} \; \text{tr} \left( \sum_{p=1}^P f_{\bbTheta_X} ( \bbx_p ) f_{\bbTheta_Y} ( \bby_p )^\Tr \right) \label{eq:dcca} \\
    \text{s. to:} \; &
    \sum_{p=1}^Pf_{\bbTheta_X} ( \bbx_p ) f_{\bbTheta_X} ( \bbx_p )^\Tr = P \bbI \nonumber \\ &\sum_{p=1}^Pf_{\bbTheta_Y} ( \bby_p )f_{\bbTheta_Y} ( \bby_p )^\Tr = P\bbI, \nonumber
\end{align}
where $f_{\bbTheta_X}$ and $f_{\bbTheta_Y}$ represent the NNs that compute the transformed views for the $\reals^{N_X}$ and $\reals^{N_Y}$ views, respectively, with parameters $\bbTheta_X$ and $\bbTheta_Y$. The approach in~\cite{andrew13dcca} is to solve \eqref{eq:dcca} by maximizing instead the trace norm (sum of singular values) of $\bbT$ and by providing the gradient of this alternative objective function with respect to the transformed views~\cite[Eqs. (11)-(13)]{andrew13dcca}.
Other works have sought to approximately solve the problem in~\eqref{eq:dcca} by implementing the following Lagrangian relaxation~\cite{chang18softCCA,zhang2021from}
\begin{align}
    \min_{\bbTheta_X, \bbTheta_Y} \; & \frac{1}{P}\sum_{p=1}^P \| f_{\bbTheta_X} ( \bbx_p ) - f_{\bbTheta_Y} ( \bby_p ) \|^2 + \label{eq:soft_dcca} \\
    &
    \begin{aligned}
    \lambda \Bigl( & \Big\| \frac{1}{P}\sum_{p=1}^Pf_{\bbTheta_X} ( \bbx_p ) f_{\bbTheta_X} ( \bbx_p )^\Tr-\bbI\Big\|_F^2 + \\
    & \Big\| \frac{1}{P}\sum_{p=1}^Pf_{\bbTheta_Y} ( \bby_p ) f_{\bbTheta_Y} ( \bby_p )^\Tr-\bbI\Big\|_F^2 \Bigr), \nonumber
    \end{aligned}
\end{align}
where rather than maximizing the correlation, the goal is to minimize the squared error. Furthermore, the two regularization terms, proposed in~\cite{chang18softCCA}, are referred to as decorrelation loss.

\section{Model Formulation}\label{S:methodology}

In this section, we present our graph-based deep learning architecture, whose schematic view is represented in Figure~\ref{fig:arch}.

First, we formally state the problem we aim to solve. Let $\ccalG_X$ be a graph with $N_X$ nodes, which we will subsequently refer to as the input graph, and let $\bbX \in \reals^{N_X \times F_X}$ be an input graph signal defined over $\ccalG_X$, where $F_X$ is the number of input features. That is, rather than considering that the input graph signal is unidimensional, we assume that every node in $\ccalG_X$ has a collection of $F_X$ values. Similarly, let $\ccalG_Y$ be an output graph with $N_Y$ nodes, and let $\bbY \in \reals^{N_Y \times F_Y}$ be the output graph signal defined over $\ccalG_Y$, with $F_Y$ output features.
The focus of this work is to find a nonlinear parametric function $f_{\bbTheta}$ that leverages the information of both $\ccalG_X$ and $\ccalG_Y$ and learns a mapping from $\reals^{N_X \times F_X}$ to $\reals^{N_Y \times F_Y}$ using a training set $\{\bbX_p,\bbY_p\}_{p=1}^P$ , with $\bbTheta$ representing the learnable parameters of the architecture.
In other words, for a given input $\bbX$ the architecture should compute its output $\hbY$ by incorporating the input signal along with information from both the input and output graphs $\hbY = f_{\bbTheta} (\bbX | \ccalG_X, \ccalG_Y)$.

A crucial aspect of the proposed architecture is the assumption that there exists a latent space $\bbZ$ where we implement the transformation from signals defined over $\ccalG_X$ to signals on $\ccalG_Y$.
Based on this assumption, $f_{\bbTheta} (\bbX | \ccalG_X, \ccalG_Y)$ can be written as the composition of three different functions (refer to Figure~\ref{fig:arch}):
\begin{enumerate}
    \item $\psi^X_{\bbTheta_X}: \reals^{N_X \times F_X} \to \reals^{N_X \times F_{Z_X}}$ is the function, parameterized by $\bbTheta_X$, that learns the mapping form the input signal $\bbX$ into the latent space $\bbZ_X \in \reals^{N_X \times F_{Z_X}}$ using $\ccalG_X$. In this work we consider a GNN with no pooling, which means that the output of $\psi^X_{\bbTheta_X}$ still has dimension $N_X$.
    \item $\psi^Z_{\bbTheta_Z}: \reals^{N_X \times F_{Z_X}} \to \reals^{N_Y \times F_{Z_Y}}$ is the function that transforms the latent space $\bbZ_X$ into $\bbZ_Y \in \reals^{N_Y \times F_{Z_Y}}$. Its parameters, if any, are contained in $\bbTheta_Z$ (a detailed discussion about this is provided in the next section). This transformation allows the architecture to start working with $\ccalG_Y$, as the dimensions of $\bbZ_Y$ match those of the output graph.
    \item $\psi^Y_{\bbTheta_Y}: \reals^{N_Y \times F_{Z_Y}} \to \reals^{N_Y \times F_Y}$ is the function, parameterized by $\bbTheta_Y$, that transforms the latent space $\bbZ_Y$ into the target signal using $\ccalG_Y$. Similarly to $\psi^X_{\bbTheta_X}$, this step can be implemented using a standard GNN.
\end{enumerate}


The overall function is therefore implemented as a composition of these three steps, i.e. $f_{\bbTheta} = \psi^Y_{\bbTheta_Y} \circ \psi^Z_{\bbTheta_Z} \circ \psi^X_{\bbTheta_X}$, where $\bbTheta = \{ \bbTheta_X, \bbTheta_Z, \bbTheta_Y\}$ contains the parameters of the whole architecture.
The output to an input $\bbX$ is therefore computed as 
\begin{equation}
\hbY = \psi^Y_{\bbTheta_Y} ( \psi^Z_{\bbTheta_Z} ( \psi^X_{\bbTheta_X} ( \bbX | \ccalG_X )) | \ccalG_Y),    
\end{equation} 
where the graph used in each function is included as prior information.

As already mentioned, our approach is to set $\psi^X_{\bbTheta_X}$ and $\psi^Y_{\bbTheta_Y}$ as GNNs with no pooling. The reason for this design decision is twofold. There are a number of available GNN architectures that are relatively well understood and have been shown to be capable of capturing the graph topology. Equally important, GNNs have been successfully deployed in a number of applications, achieving state of the art results. The design of  $\psi^Z_{\Theta_Z}$, the last block in our architecture, is addressed in Section \ref{SS:transformation} but before a closing comment regarding the training of the architecture is due. 

\vspace{.1cm}
\noindent \textbf{Training of the architecture:} Training is run as in many other graph-based machine learning models. Specifically, if we are dealing with a \textit{fully supervised setup}, we are given $\{\bbX_p,\bbY_p\}_{p=1}^P$ along with $\ccalG_X$ and $\ccalG_Y$. In that case, the parameters are learned to minimize 
\begin{equation}
    \hbTheta =\underset{\bbTheta_X, \bbTheta_Z, \bbTheta_Y}{\arg\min}\sum_{p=1}^P \ccalL\big(\bbY_p , \hbY_p \big),
\end{equation}
where $\hbY_p=\psi^Y_{\bbTheta_Y} ( \psi^Z_{\bbTheta_Z} ( \psi^X_{\bbTheta_X} ( \bbX_p | \ccalG_X )) | \ccalG_Y)$ and the loss function $\ccalL$ depends on the nature of the data in $\bbY$ (categorical, numeric, Boolean...). In many practical cases, the loss will be written as a sum over the different entries of $\bbY_p$ (i.e., nodes and features). 

In some other scenarios, we may have access only to a single $\bbX_p$ and a few nodes of $\ccalG_Y$ so that only some of the rows of $\bbY_p$ are available for training. Let $\ccalV_Y^{tr} \subset \ccalV_Y$ be the subset of the nodes in $\ccalG_Y$ where the output is observed. Then, in this \textit{semi-supervised setting}, the parameters are learned to minimize 
\begin{equation}
    \hbTheta =\underset{\bbTheta_X, \bbTheta_Z, \bbTheta_Y}{\arg\min}\sum_{n\in\ccalV_Y^{tr}} \ccalL_n\big([\bbY_p]_{n,:} ,[\hbY_p]_{n,:}\big),
\end{equation}
where $[\bbY_p]_{n,:}$ ($[\hbY_p]_{n,:}$) denotes the $n$-th row of $\bbY_p$ ($\hbY_p$) and $\ccalL_n$ is the per-node loss function, which depends on the nature of the features in $[\bbY_p]_{n,:}$. 

\subsection[Implementation of $\psi^Z_{\Theta_Z}$]{Implementation of $\psi^Z_{\boldsymbol{\Theta}_Z}$} \label{SS:transformation}

\begin{figure*}
    \centering
    \includegraphics[width=\textwidth]{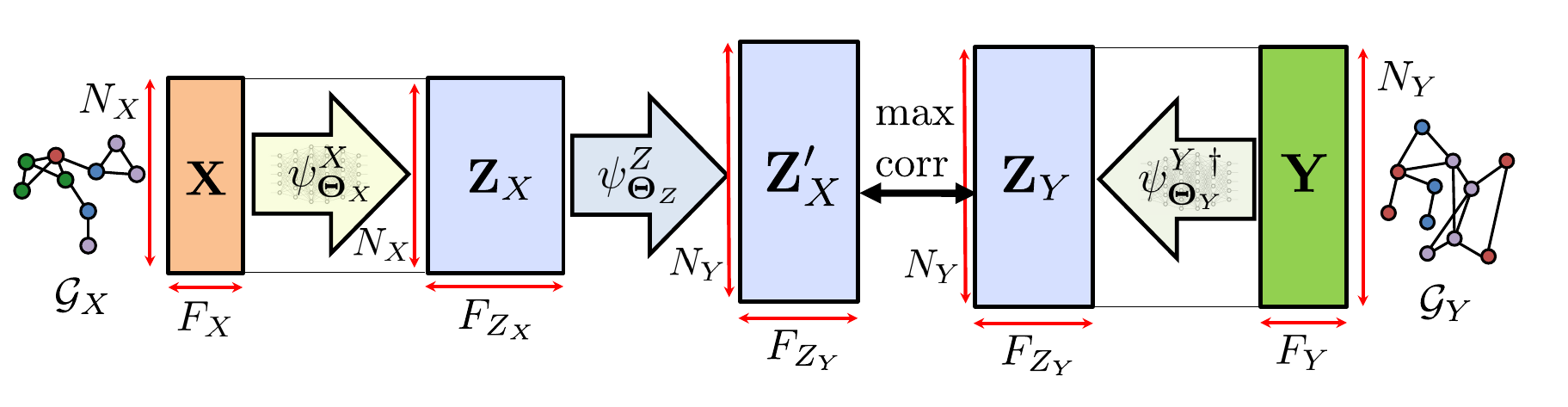}
    \caption{Modification of the architecture to find alternative representations of our data and possibly use them in downstream tasks. Here, we don't focus on the output space, but on the latent space instead, aiming to maximize the correlation between the transformed features $\bbZ_X'$ and $\bbZ_Y$ (we cannot use $\bbZ_X$ directly as it is defined on a different graph than $\bbZ_Y$). The learned representations can then be used in downstream tasks.}
    \label{fig:arch_cca}
\end{figure*}

The role of $\psi^Z_{\bbTheta_Z}$ is to apply a transformation in the latent space that maps a collection of signals (features) of dimension $N_X$ to a collection of signals of dimension $N_Y$. Different alternatives for this transformation exist. Rather than focusing on a particular one, this section aims to provide an understanding of different meaningful approaches to designing this function.

To better understand the trade offs involved, when designing $\psi^Z_{\bbTheta_Z}$ one should take into account if this function is:
\begin{itemize}
    \item \textbf{Domain specific or agnostic to the underlying task}: When domain knowledge of the underlying scenario represented by our graphs is available, we may be able to design a domain-specific transformation that leverages this knowledge.
    For example, if we know in advance that $\ccalG_X$ is a coarsened version of $\ccalG_Y$ (or vice versa), we may choose to copy or aggregate the embeddings from the nodes based on their membership.
    If this possibility is not present, resorting to a generic transformations is more prudent. Some examples of domain-agnostic transformations are provided next, while the possibility of designing a domain-specific transformation will be explored in Section~\ref{S:experiments}.
    \item \textbf{Learnable vs fixed in advance}: The transformation function $\psi^Z_{\bbTheta_Z}$ is the intermediary step of two GNN models. One alternative is to fix the transformation beforehand, e.g., using an interpolation or sampling operator between the two graphs. When no such prior knowledge exists, it is reasonable to propose a parametric function and learn its parameters jointly with the parameters of the GNN via GD. 
    \item \textbf{Linear vs more complex transformations}: Linear transformations are an easy-to-implement versatile choice. An important point here is that, since we are working with multidimensional signals, the most general case entails working with the vectorized version of the signals in the latent space so that we have
    \begin{equation}
        \text{vec}(\bbZ_Y) = \bbW \text{vec}(\bbZ_X),
        \label{eq:linear_psi_Z}
    \end{equation}
    where $\bbW \in \reals^{N_Y F_{Z_Y} \times N_X F_{Z_X}}$ is a (possibly learnable) transformation matrix.
    The number of elements in $\bbW$ depends partly on the number of nodes in the input and output graphs $N_X$ and $N_Y$. If the graphs are large (as it happens in most real world networks), it renders the number of elements in $\bbW$ very high. This leads to two issues, (i) the matrix multiplication increases in complexity with the cube of the number of elements in the matrix, rendering the operation computationally expensive and (ii) if this matrix is learned through backpropagation, the very high number of learnable parameters can lead to overfitting.
    An effective way to deal with this issue is to incorporate some structure onto $\bbW$. Relevant examples include low-rank, Kronecker-structured and permutation matrices. In the first case, we simply consider the low-rank matrix $\bbW = \bbW_Y \bbW_X^\Tr$, where $\bbW_Y \in \reals^{N_Y F_{Z_Y} \times K}$ and $\bbW_X \in \reals^{N_X F_{Z_X} \times K}$ define the transformation with a lower number of parameters as $K \ll N_Y F_{Z_Y}$ and $K \ll N_X F_{Z_X}$. If both matrices are learnable, this bilinear low-rank representation reduces the number of parameters from $N_Y F_{Z_Y} N_X F_{Z_X}$ to $(N_Y F_{Z_Y} + N_X F_{Z_X}) K$. Both $\bbW_X$ and $\bbW_Y$ could be learned jointly via backpropagation or in an alternating fashion. Clearly, one: (i) could fix one of the matrices and learn only the second one, or (ii) consider multiplications of more than 2 matrices. 
    
    Another practical option to reduce the degrees of freedom is to endow $\bbW$ with a Kronecker structure, either by considering $\bbW = \bbW_F \otimes \bbW_N$, which would be equivalent to compute the transformation in $\psi^Z_{\bbTheta_Z}$ as $\bbZ_Y = \bbW_N \bbZ_X \bbW_F^\Tr$\footnote{Please note the property $\text{vec} (\bbA \bbB \bbC) = (\bbC^T \otimes \bbA) \text{vec}(\bbB)$.}, or by considering $\bbW = \bbW_F \oplus \bbW_N$, which would be equivalent to consider $\bbZ_Y = \bbW_N \bbZ_X + \bbZ_X \bbW_F^\Tr$. The matrix $\bbW_N \in \reals^{N_Y \times N_X}$ combines the information across nodes and $\bbW_F \in \reals^{F_{Z_Y} \times F_{Z_X}}$ does the same across features. Both $\bbW_N$ and $\bbW_F$ can be fixed in advance (e.g. either using domain knowledge or random sketching matrices) or learned jointly with the rest of the parameters. In the latter case, the number of degrees of freedom of the associated (bilinear) mapping $N_Y N_X + F_{Z_Y} F_{Z_X}$.
    
    A last example to incorporate structure onto the linear transformation is to implement $\bbW$ as a permutation matrix, so that the cells in $\bbZ_Y$ are a rearranged version of the cells in $\bbZ_X$.
    
    The implementation of $\psi^Z_{\bbTheta_Z}$ using a (bi-)linear function provides for some insights and interpretability, at the potential cost of sacrificing expressibility. Clearly, \textit{learnable nonlinear} mappings can be considered as well. In the context of NNs, multiple $\psi^Z_{\bbTheta_Z}$ are well motivated including (i) a feedforward neural network with a learnable $N_Y F_{Z_Y} \times N_X F_{Z_X}$ matrix, (ii) a Two-Layer Perceptron with two learnable $N_Y F_{Z_Y} \times K$ and $K \times N_X F_{Z_X}$ matrices; and (iii) a Multi-Layer Perceptron (MLP) which operates only across rows and whose first and last layers have dimensions $N_X$ and $N_Y$, respectively. In all those cases, the approach is to learn the parameters of the NN implementing $\psi^Z_{\bbTheta_Z}$ are learned jointly with the parameters of the input and the output GNNs.
\end{itemize}


We close this section by specifying two examples of transformations that (i) are easy to implement and (ii) will be used in the numerical experiments. Both of them are non-domain specific linear transformations, hence they can be employed even without prior information:

\begin{itemize}
    \item $\bbZ_Y = \bbZ_X^\Tr$: This non-learnable transformation, which is essentially a permutation operator, sets $F_{Z_X} = N_Y$ and $F_{Z_Y} = N_X$. This should be considered when designing the number of features in the output and input layers of $\psi^X_{\bbTheta_X}$ and $\psi^Y_{\bbTheta_Y}$, respectively. Under this simple approach, the features learned by the input GNN are associated with the nodes of the second GNN, and vice-versa. 
    \item $\bbZ_Y = \bbW_N \bbZ_X$: In this case, we set $F_{Z_X} = F_{Z_Y}$ and $\bbW_N \in \reals^{N_Y \times N_X}$ is learned jointly with the rest of the parameters of the architecture via backpropagation. This is a learnable linear transformation with Kronecker structure, where $\bbW$ in  \eqref{eq:linear_psi_Z} is merely $\bbW = \bbI \otimes \bbW_N$.
\end{itemize}

Clearly, there are plenty of options to implement $\psi^Z_{\bbTheta_Z}$. The particular choice should depend on aspects such as the number of training samples, the application domain, or the computational complexity that can be afforded, among others. In the numerical experiments section, the two example alternatives provided here, as well as some domain-specific transformations, will be explored and its performance will be evaluated.

\bigskip

\vspace{0.2cm}
\noindent
\textbf{Remark 1}:
It is worth noting the flexibility of the proposed architecture as (i) the GNN used to implement both $\psi^X_{\bbTheta_X}$ and $\psi^Y_{\bbTheta_Y}$ is not fixed (any architecture whose input and output are signals defined over the same graph, such as the GCN~\cite{kipf2017semisupervised} or GAT~\cite{velickovic2018graph} architectures, can be used), and (ii) the function used to implement $\psi^Z_{\bbTheta_Z}$ is also chosen as the best option considering the underlying task we want to solve.

\vspace{0.2cm}
\noindent
\textbf{Remark 2}:
The parameters $\bbTheta$ of $f_{\bbTheta}$, namely the learnable parameters of the functions $\psi^X_{\bbTheta_X}$, $\psi^Y_{\bbTheta_Y}$ and (possibly) $\psi^Z_{\bbTheta_Z}$, are learned jointly via GD and backpropagation as long as $\psi^Z_{\bbTheta_Z}$ is differentiable. Note that, for example, every (linear) transformation defined by~\eqref{eq:linear_psi_Z} is differentiable and thus allows the gradient calculation and backpropagation, and the same holds true for non-linear differentiable functions such as the MLP.

\section{Connection with Canonical Correlation Analysis and Self-Supervised Learning} \label{S:cca_ssl}

Until now, our focus has been on the output space and the output graph. However, if we shift our attention to the underlying latent space, similarities with other techniques, such as CCA and SSL, emerge. In this section, we aim to find a latent space that contains information about both the input and output graphs, and possibly use it for a downstream task.

As explained in Section \ref{S:fundamentals}, given a pair of views $(\bbx_p,\bby_p)$ of the same (related) object, CCA projects the two views onto a common low dimensional space using linear orthogonal transformations $\bbU$ and $\bbV$, so that $\bbz_{X,p}=\bbU\bbx_p$ and $\bbz_{Y,p}=\bbV\bby_p$. The transformations $\bbU$ and $\bbV$ are designed to maximize the correlation between the transformed views $\bbz_{X,p}$ and $\bbz_{Y,p}$. Utilizing our framework, an analogous process can be implemented, replacing $\bbU$ and $\bbV$ with graph-aware NN-based transformations.

More formally, given a training set $\{(\bbX_p,\bbY_p)\}_{p=1}^P$ of \textit{multidimensional} graph-signals, together with the graphs $\ccalG_X$ and $\ccalG_Y$, the goal here is to find transformed views of our data $\bbZ_{X,p}$ and $\bbZ_{Y,p}$ such that they are maximally correlated. While in a purely unsupervised case the reason for finding these lower-dimensional representations can be simply compression or denoising, in many setups the transformed views $\bbZ_{X,p}$ and $\bbZ_{Y,p}$ are employed to solve a downstream task. This includes the case where the goal is to estimate a third signal $\bbrho$ defined over either $\ccalG_X$ or $\ccalG_Y$, which is a SSL setup that will be discussed in the next subsection.

Regardless of the final application, our approach for creating the low-dimensional representations is to adopt the formulation in \eqref{eq:soft_dcca}, adapting it to the setup at hand\footnote{If convenient, counterpart graph deep CCA formulations based on \eqref{eq:dcca} can be posited as well.}. To find the views, we first implement the GNNs over $\ccalG_X$ and $\ccalG_Y$, yielding the multidimensional signals $\bbZ_X = \psi^X_{\bbTheta_X} (\bbX | \ccalG_X)$ and $\bbZ_Y = \psi^{Y \; \dagger}_{\bbTheta_Y} (\bbY | \ccalG_Y)$. Notice that, to maintain notational consistency with the previous section, we denote the GNN operating on the view $\bbY$ as $\psi^{Y \; \dagger}_{\bbTheta_Y}$. Since the dimensions of $\bbZ_X$ and $\bbZ_Y$ do not match, we further apply the function $\psi^Z_{\bbTheta_Z}$ to $\bbZ_X$. This yields $\bbZ_X' = \psi^Z_{\bbTheta_Z}(\bbZ_X)$, which has the same dimensions as $\bbZ_Y$. Alternatively, we can define the transformation $\psi^{Z \; \dagger}_{\bbTheta_Z}$, apply it to $\bbZ_Y$ and obtain the representation $\bbZ_Y' = \psi^{Z \; \dagger}_{\bbTheta_Z} (\bbZ_Y)$, which has the same dimensions as $\bbZ_X$. In practice, a reasonable approach is to determine $\max (N_X, N_Y)$ and apply $\psi^Z_{\bbTheta_Z}$ to the matrix with a larger size, so that the latent representation lives in a lower dimensional space. 
Assume, without loss of generality, that $N_X=\max (N_X, N_Y)$, we then propose solving the following optimization problem
%
%
%
\begin{align}
        &\hbTheta \!=\!\!\! \underset{\bbTheta_X, \bbTheta_Z, \bbTheta_Y}{\arg\min} \frac{1}{P}\! \sum_{p=1}^P \!\|  \psi^Z_{\bbTheta_Z}\! ( \psi^X_{\bbTheta_X} ( \bbX_p | \ccalG_X )) \!-\! \psi^{Y \; \dagger}_{\bbTheta_Y} ( \bbY_p | \ccalG_Y) \|_F^2 \nonumber\\
    &\!+\!\lambda \Big(  \Big\| \frac{1}{P} \!\sum_{p=1}^P \!\psi^Z_{\bbTheta_Z} ( \psi^X_{\bbTheta_X} ( \bbX_p | \ccalG_X ))^\Tr \psi^Z_{\bbTheta_Z} ( \psi^X_{\bbTheta_X} \!( \bbX_p | \ccalG_X )) \! - \! \bbI \Big\|_F^2 \nonumber \\
    & \quad +\Big\| \frac{1}{P}\sum_{p=1}^P\psi^{Y \; \dagger}_{\bbTheta_Y} ( \bbY_p | \ccalG_Y)^\Tr \psi^{Y \; \dagger}_{\bbTheta_Y} ( \bbY_p | \ccalG_Y) - \bbI \Big\|_F^2\Big),\label{eq:cca_problemv0}
\end{align}
where we recall that $\bbTheta = \{\bbTheta_X, \bbTheta_Z, \bbTheta_Y\}$ collects the learnable parameters of the architecture. A schematic view of the proposed approach, with $\psi^Z_{\bbTheta_Z}$ applied over $\bbZ_X$, is shown in Figure~\ref{fig:arch_cca}.
It is important to note that the above minimization promotes the orthonormality of the columns of $\bbZ_X$ and $\bbZ_Y$. In any case, the differentiability of the different terms and the absence of constraints facilitates the optimization, enabling the use of a simple stochastic gradient descent approach and yielding superior results in experiments related to SSL and downstream tasks.

As in Section \ref{S:methodology}, there may be setups were we do not have access to a training set with multiple samples, but we have access only to a single observation pair. In that case, the learning is carried out simply by setting $P=1$, so that  
\begin{align}\label{eq:cca_problem}
    &\hbTheta =\!\underset{\bbTheta_X, \bbTheta_Z, \bbTheta_Y}{\arg\min} \| \psi^Z_{\bbTheta_Z} ( \psi^X_{\bbTheta_X} ( \bbX_p | \ccalG_X )) - \psi^{Y \; \dagger}_{\bbTheta_Y} ( \bbY_p | \ccalG_Y) \|_F^2   \nonumber\\
    & +\lambda \Big( \| \psi^Z_{\bbTheta_Z} ( \psi^X_{\bbTheta_X} ( \bbX_p | \ccalG_X ))^\Tr \psi^Z_{\bbTheta_Z} ( \psi^X_{\bbTheta_X} ( \bbX_p | \ccalG_X )) - \bbI \|_F^2  \nonumber \\
    &\quad\quad + \| \psi^{Y \; \dagger}_{\bbTheta_Y} ( \bbY_p | \ccalG_Y)^\Tr \psi^{Y \; \dagger}_{\bbTheta_Y} ( \bbY_p | \ccalG_Y) - \bbI \|_F^2\Big). 
\end{align}
In such setups, it is prudent to postulate architectures with a reduced number of parameters, so that learning can be effectively accomplished based on the number of training values.

Finally, it is worth mentioning that an equally valid approach for computing the views is to consider two different functions in the latent space: $\psi^{Z_X}_{\bbTheta_{Z_X}}$, which is applied to $\bbZ_X$, and $\psi^{Z_Y}_{\bbTheta_{Z_Y}}$, which is applied to $\bbZ_Y$. This would yield a symmetric architecture and further reduce the dimensionality of the transformed views. This, for example, can be accomplished by using two MLPs or, more simply, by implementing the factorizable low-rank transformation $\bbW = \bbW_Y \bbW_X^\Tr$ introduced in Section~\ref{SS:transformation}. In the latter setting, $\bbZ_X$ and $\bbZ_Y$ are mapped to a common domain of dimensionality $K$ just by multiplying them by $\bbW_X^\Tr$ and $\bbW_Y^\Tr$, respectively.

\subsection{Self-supervised learning} \label{SS:ssl}

The transformed features $\bbZ_X' = \psi^Z_{\bbTheta_Z} ( \psi^X_{\bbTheta_X} ( \bbX | \ccalG_X ))$ and $\bbZ_Y = \psi^{Y \; \dagger}_{\bbTheta_Y} ( \bbY | \ccalG_Y)$ encapsulate information from both the input and output views, namely $\bbX$ and $\bbY$, as well as the input and output graphs, denoted as $\ccalG_X$ and $\ccalG_Y$. As previously mentioned, these transformed features can be used for compression or denoising tasks. However, the absence of label information in training the GNNs for the CCA setting aligns with the domain of SSL as well. Within this domain, we are in position to train a simpler classifier $g_{\bbXi}(\cdot): \reals^{F_{Z_Y}} \rightarrow \reals$ with parameters $\bbXi$, such as a logistic regressor, on these transformed features, which allows us to address a downstream task by using a small amount of labels. In this subsection we explore this setting.

To exemplify the approach more formally, let us suppose that: (i) our interest is on the labels (values) of signal $\bbrho \in \reals^{N_Y}$; and (ii) we have access to the labels only for the subset of nodes $\ccalV_Y^{tr} \subset \ccalV_Y$, with $|\ccalV_Y^{tr}| \ll N_Y$.
Moreover, let $[\bbZ_X']_{n,:} \in \reals^{F_{Z_Y}}$ be the $n$-th row of $\bbZ_X'$.
We then consider the problem 
\begin{equation}
    \hbXi=\arg\min_{\bbXi} \sum_{n \in \ccalV_Y^{tr}} \ccalL_{SSL} (g_{\bbXi} ([\bbZ_X']_{n,:}), \rho_n),
\end{equation}
where $\ccalL_{SSL}$ represents a suitable loss function for the SSL task. 
After obtaining the parameters $\hbXi$, we can use the trained classifier $g_{\hbXi}$ to obtain the labels in the remaining nodes $\ccalV_Y \backslash \ccalV_Y^{tr}$ using as input the lower-dimensional representations $[\bbZ_X']_{n,:}$ for all $n\in \ccalV_Y \backslash \ccalV_Y^{tr}$. 
Note that, different from the semi-supervised setting, we don't use the whole graph signal $\bbZ_X'$, as the row $[\bbZ_X']_{n,:}$ corresponding to the node already contains the information of both the input and output graphs, as well as the views supported in each of them. 


In a nutshell, the setting proposed using the CCA-based problem in~\eqref{eq:cca_problem} allows to learn informative transformed views that encompass information from both the input and output graph, and the associated signals in both of them, without using label information. Then, the motivation to train a simpler classifier is that, using a small number of labeled samples, we are able to obtain a performance close to a supervised or semi-supervised setting with a larger training set, due to the quality of the transformed views. Considering the difficulty that lies in labeling samples, this may allow us to learn in more challenging scenarios where the size of the training set is small.
An illustrative example of this application will be provided in Section~\ref{SS:ccassl}.

\section{Numerical Results} \label{S:experiments}

In this section, we empirically evaluate the performance of the proposed architecture in different scenarios. Rather than exhaustively testing all the proposed alternatives, our goal is to illustrate the relatively broad applicability of the multi-graph architecture. The code repository with all the details of the implementation, additional test cases, and additional evaluation metrics is available at: \url{https://github.com/vmtenorio/io-gnn}.

\subsection[Baselines and choices for $\psi^Z_{\Theta_Z}$]{Baselines and choices for $\psi^Z_{\boldsymbol{\Theta}_Z}$} \label{SS:exp_setup}

Our architecture involves 3 blocks. For the first block $\psi^X_{\bbTheta_X}$ and third block $\psi^Y_{\bbTheta_Y}$ we consider a GCN, a GAT, and an MLP. These design choices are labelled as ``IOGCN'', ``IOGAT'' and ``IOMLP'', respectively. When comparing their performance, the experiments are carried out using the same number of layers, number of neurons and number of hidden features, and training them with the same splits and algorithms, so that the comparisons are ``fair''. 
When possible, we will also implement a GCN that does not leverage the information in the input and output graphs but, rather, operates using only one graph. This setting differs among experiments, so the specific setup will be explained in each of the sections.

The choices for the transformation function $\psi^Z_{\bbTheta_Z}$ will be specified in each test case, but we mostly focus on the two examples provided in Section~\ref{SS:transformation}, namely the transpose permutation (``-T'') and the learnable linear mapping (``-W''), as well as on domain-specific transformations that will be detailed in each section.

\subsection{Subgraph feature estimation} \label{SS:subgraph}

In the first experiment, we utilize the Cora graph dataset~\cite{mccallum2000automating}, which is a commonly used citation network. We create two subgraphs, namely $\ccalG_X$ and $\ccalG_Y$, from the original network $\ccalG$. Specifically, we randomly select a node from $\ccalG$ and compute the subgraph encompassing nodes within the $k$-hop neighborhood of the chosen node to generate both $\ccalG_X$ and $\ccalG_Y$.
More details about this graph sampling method, known as snowball sampling, can be found in, e.g.,~\cite{stivala16snowball}.
The primary objective is to assess the learning capabilities of the architecture when the input signal is known for a subset of nodes, but the goal is to predict the output in a different set, which may have overlapping elements.
In alignment with the original Cora's node classification task, we consider a semi-supervised setting. The nodes for the training, validation, and test sets are uniformly chosen at random from the nodes in $\ccalG_Y$, with the partition sizes set to 30\%, 20\%, and 50\% of the nodes, respectively.
To mitigate the impact of random factors such as the choice of the initial node, GNN parameter initialization, and the partitioning of training, validation, and test sets, we perform the experiments 25 times. The reported results represent the mean values across these 25 realizations. As stated earlier, additional metrics can be found in the code repository.

Once we have defined how the input and output graphs are created and provided the details of the experiment, the next step is to describe the form of $\psi^Z_{\bbTheta_Z}$, which, for this experiment, is a domain-specific transformation. Specifically, since we know the overlapping nodes in $\ccalG_X$ and $\ccalG_Y$, the ``common'' transformation for $\psi^Z_{\bbTheta_Z}$ copies the values from the rows of $\bbZ_X$ to those of $\bbZ_Y$ corresponding to the overlapping nodes between the two graphs. The rest of the rows of $\bbZ_Y$, corresponding to the nodes in $\ccalG_Y$ not present in $\ccalG_X$, are zero-padded. This domain-specific transformation will be labelled as ``-C'' in the legends.

\begin{figure}
    \centering
    \includegraphics[width=0.5\textwidth]{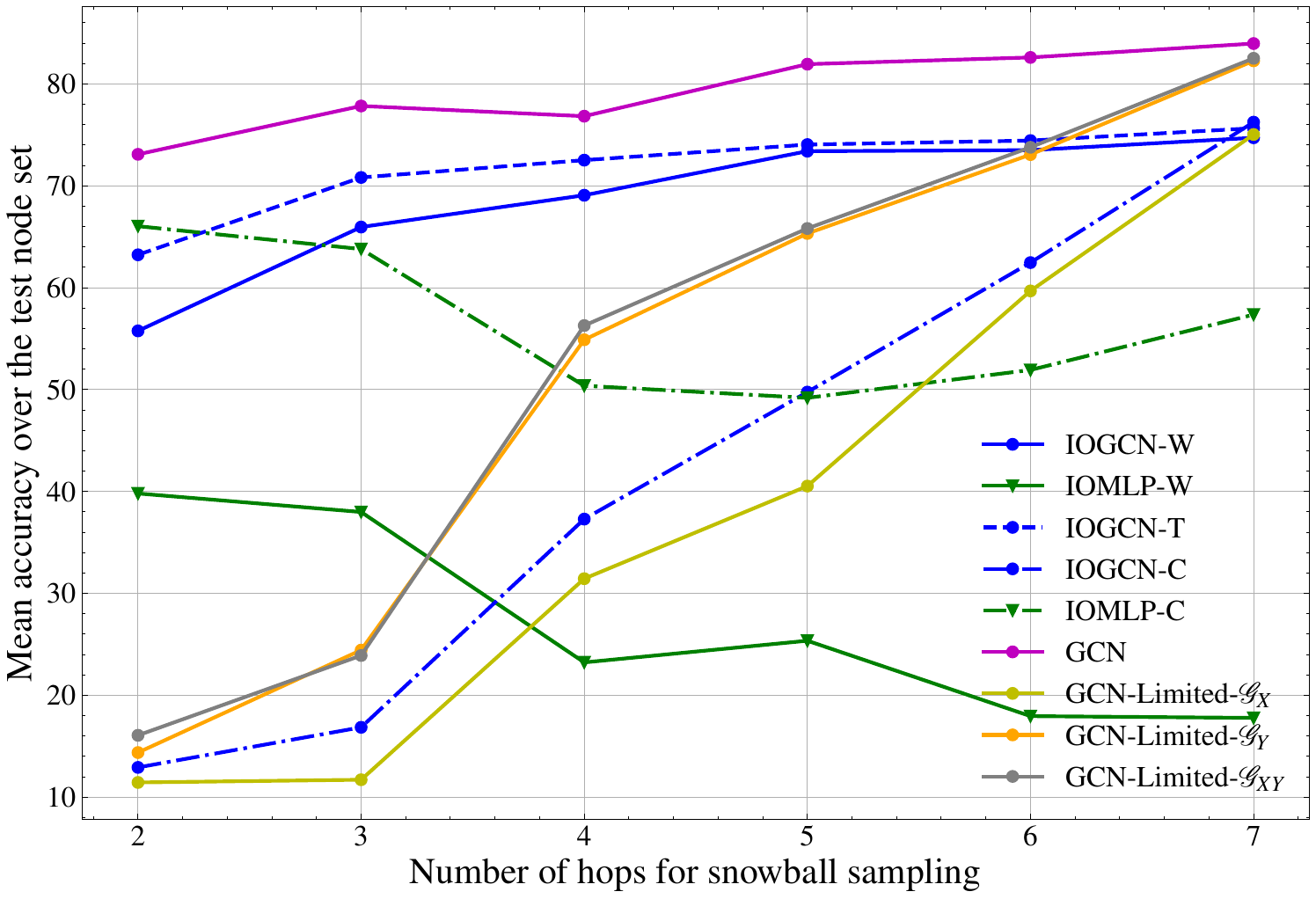}
    \caption{Node classification accuracy for the Cora graph dataset using two subgraphs. The subgraphs $\ccalG_X$ and $\ccalG_Y$ are drawn randomly by selecting one root node for each of them and, then, implementing a snowball sampling technique. The x-axis indicates the number of hops considered in the snowball sampling scheme used to build the graphs. The y-axis reports the classification accuracy  measured over the test node set (subset of the nodes of $\ccalG_Y$) averaged over 25 different graph realizations. The different lines correspond to the schemes introduced in Sections \ref{SS:exp_setup} and \ref{SS:subgraph}.}
    \label{fig:subgraph}
\end{figure}

Regarding baselines, we also report the results of a GCN architecture (labelled as ``GCN'') that uses the whole graph $\ccalG$ and therefore has more information available than IOGCN and IOMLP architectures, while still being evaluated on the nodes from $\ccalG_Y$ belonging to the randomly sampled test set previously defined.
Additionally, to simulate the performance of a GCN with the same information as our architecture, we construct a graph where every edge is dropped and every feature is masked except those among the subset of nodes of $\ccalG_X$ (``GCN-Limited-$\ccalG_X$''), $\ccalG_Y$ (``GCN-Limited-$\ccalG_Y$'') or the union of both (``GCN-Limited-$\ccalG_{XY}$'').

In Figure~\ref{fig:subgraph}, the performance of the architectures is depicted as the sizes of the neighborhood used to generate $\ccalG_X$ and $\ccalG_Y$ increase. Clearly, as $\ccalG_X$ and $\ccalG_Y$ encompass more nodes, they become more similar to each other (they contain more overlapping nodes). This explains the improvement in performance of the GCN operating with limited information and the IOGCN that uses the domain-specific transformation leveraging overlapping nodes.
Nonetheless, the proposed framework, both using a GCN and a GAT, is able to maintain good performance even when the sizes of $\ccalG_X$ and $\ccalG_Y$ are small, and thus the two graphs are very different. The proposed framework nearly matches the performance of the GCN operating with full information and notably outperforms IOMLP.
Regarding the differences in performance due to the different implementations of $\psi^Z_{\bbTheta_Z}$, we see that the best performing transformation, at least for smaller $\ccalG_X$ and $\ccalG_Y$, is either the transpose or the learnable linear mapping in the IOGCN approach. However, in the IOMLP, using the graph structure in the form of the overlapping nodes yields a higher accuracy, most probably due to the fact that the MLP does not consider the graph structure in either $\psi^X_{\bbTheta_X}$ nor $\psi^Y_{\bbTheta_Y}$.

\subsection{Image interpolation and segmentation} \label{SS:imageintr}

\begin{figure*}
    \centering
    \includegraphics[width=1\textwidth]{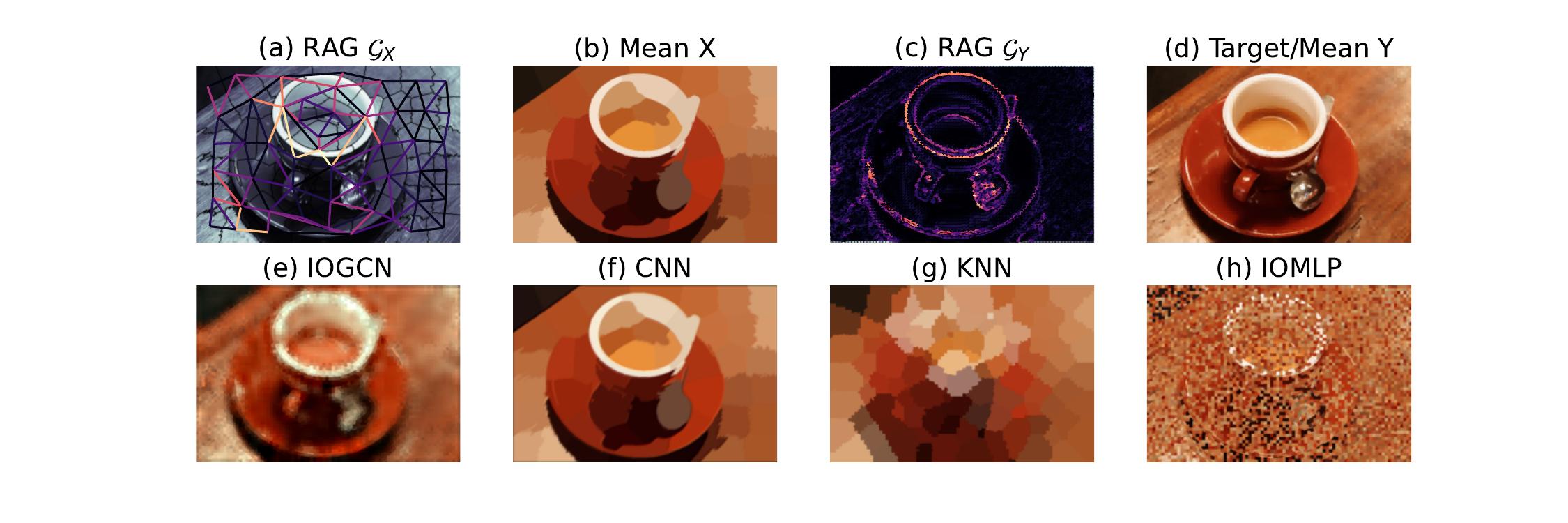}
    \caption{Sample images obtained in the image interpolation experiment, for the ``coffee'' image. The first row represents the input and output graphs $\ccalG_X$ (a) and $\ccalG_Y$ (c) and the input and output signals $\bbX$ (b) and $\bbY$ (d). The second row represents the output of the GNN-based architecture proposed, IOGCN (e), and the alternative baselines, CNN (f), KNN (g) and IOMLP (h).}
    \label{fig:sample_img}
\end{figure*}

In the second and third experiments, we assess the potential of our scheme in the context of interpolation. Here, the input graph serves as a coarse representation of an object, with the input signal representing a specific feature of interest associated with the nodes of the graph. The objective is to generate a finer representation of the same signal, defined on an output graph that offers a more detailed depiction of the same object.
The finer the representation, the more detailed the insight into the underlying signal of interest.
Naturally, two graphs emerge, and the aim is to map the signal from one graph (the coarser one) to the other (the finer one).
Two potential applications of the proposed framework for the interpolation task are explored next: image segmentation, which is the focus of this section, and fluid flow prediction, which is the focus of the next section.

In our first interpolation example, we work with images from the \verb|skimage| Python package that are divided into regions, often referred to as superpixels, using an image segmentation algorithm such as SLIC~\cite{achanta12slic}.
Both the input and output graphs are constructed as the Region Adjacency Graph (RAG) of these superpixels, where each superpixel is a node and two nodes are connected if they are adjacent in the image. The objective is to predict the colors in a finer representation of the image (smaller superpixels), given the values of the colors in a coarse representation (larger superpixels).

More formally, let $\ccalX \in \reals^{W \times H \times 3}$ denote the tensor containing the pixels in the image of size $W \times H$ with three channels: Red, Green and Blue (RGB). We apply the segmentation algorithm to transform the original image into an irregular structure formed by superpixels, and set the colors in each superpixel as the average of the colors of the pixels inside it.
This operation is represented by the function $f_{N_X} : \reals^{W \times H \times 3} \to \reals^{{N_X} \times 3}$ where $N_X$ is the target number of superpixels. The superpixel data is computed as $f_{N_X} (\ccalX) \in \reals^{{N_X} \times 3}$.
Let $\ccalV_X$ and $\ccalV_Y$ be the node sets of the input (coarse) and output (fine) graphs, $\ccalG_X$ and $\ccalG_Y$, respectively, with $|\ccalV_X| = N_X$ and $|\ccalV_Y| = N_Y$.
The goal of our architecture is, given $\ccalG_X$, $\ccalG_Y$ and the input signal $\bbX = f_{N_X} (\ccalX)$ defined over $\ccalG_X$, to compute the signal $\bbY = f_{N_Y} (\ccalX)$ defined over $\ccalG_Y$ using the proposed Input-Output framework.
For future reference, let $f^\dagger : \reals^{{N_X} \times 3} \to \reals^{W \times H \times 3}$ be a function that returns from the superpixel domain to the original image domain by assigning to each pixel in the original image the color value of the superpixel to which it belongs.
The interpolation task is treated as a semi-supervised learning task where we assume knowledge of the signal in a subset of nodes of the output graph $\ccalV_Y^{tr} \subset \ccalV_Y$.
Our focus is to predict the colors of the remaining superpixels $\ccalV_Y \backslash \ccalV_Y^{tr}$.
The superpixels in $\ccalV_Y^{tr}$ are selected uniformly at random from the superpixels in $\ccalV_Y$, and the simulations are repeated 10 times to obtain robust results, regardless of the superpixels in $\ccalV_Y^{tr}$.

The first row of Figure~\ref{fig:sample_img} shows an example of the input and target output. The images labelled as RAG $\ccalG_X$ and RAG $\ccalG_Y$ represent the input and output graphs, respectively; the image labeled ``Mean X'' represents the input signal $\bbX = f_{N_X} (\ccalX)$ (to obtain its representation in the original image size we compute $f^\dagger(\bbX)$), and ``Target/Mean Y'' represents the output/target signal $\bbY = f_{N_Y} (\ccalX)$. Since this is an interpolation task, $N_Y > N_X$.

Once we have described the problem to solve, the next steps are to detail the transformation used in $\psi^Z_{\bbTheta_Z}$ for this setting and the baselines used for the comparison.
Regarding the former point, in this experiment we consider a domain-specific transformation that assigns to each node in $\ccalG_Y$ the average of the signal $\bbZ_X$ in the $k=2$ closest nodes from $\ccalG_X$, where the distance between nodes is measured in terms of the distance between the centers of each superpixel.
We will refer to this approach as ``selection'', as we are selecting the nodes in $\ccalG_X$ from which to copy the information to the nodes of $\ccalG_Y$, and represented as ``-S'' in the figure legends.
We compare the performance of our architecture IOGCN with IOMLP and two baselines specific for this experiment: (i) a Convolutional Neural Network (CNN) that takes as input the image resulting from the operation $f^\dagger(f_{N_X} (\ccalX))$ and does not include pooling, along with the necessary padding and stride to ensure that the output is the same size as the input and (ii) a $k$-Nearest Neighbors (KNN) approach where the selection-based transformation $\psi^Z_{\bbTheta_Z}$ is directly applied to the input, thereby predicting $\hbY = \psi^Z_{\bbTheta_Z} (\bbX)$.

All methods are used to obtain a prediction $\hbY \in \reals^{N_Y \times 3}$ defined over the output graph $\ccalG_Y$, which is represented as an image by computing $f^\dagger (\hbY)$. The results are depicted in the second row of Figure~\ref{fig:sample_img}, where we qualitatively observe that our proposed architecture, IOGCN, is able to better interpolate the color values of the superpixels in the output graph, learning some features of the original image not present in the input, such as the reflections on the spoon.
Regarding the alternatives, CNN merely copies the input image, KNN yields a poor result, and IOMLP clearly overfits to the training pixels while struggling to interpolate in the rest.

While Figure~\ref{fig:sample_img} provides qualitative results that allow us to visualize the recovered images, we also quantify the error between the colors obtained by our framework and the colors of the pixels in the original image in a more systematic way by computing the Mean Squared Error (MSE) as
%
\begin{equation}
    \frac{1}{3WH} \sum_{i=1}^W \sum_{j=1}^H \sum_{c=1}^3 [f^{\dagger} (\hbY) - \ccalX]_{ijc}^2.
    \label{eq:mse}
\end{equation}

\begin{figure}
    \centering
    \includegraphics[width=0.5\textwidth]{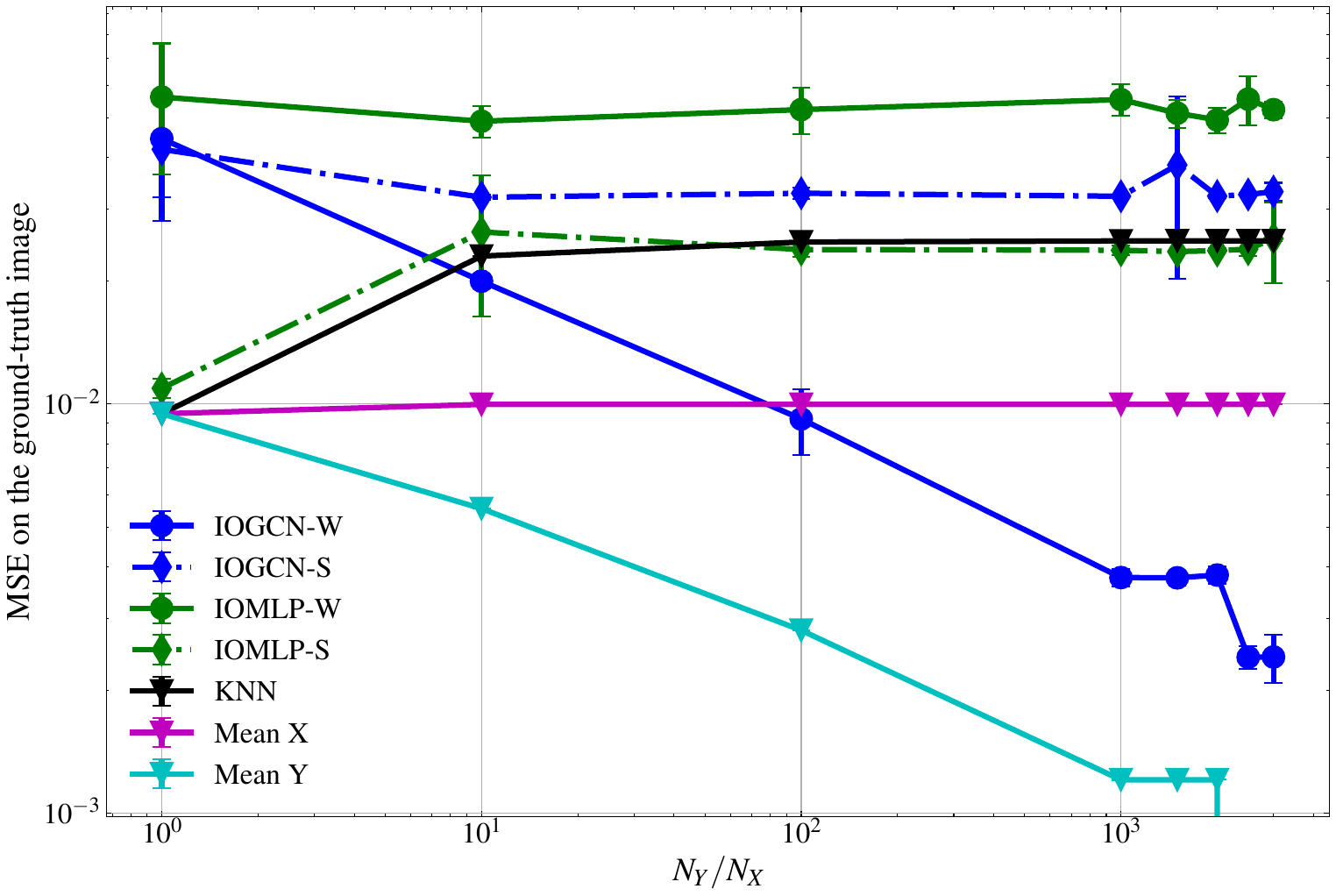}
    \caption{Error obtained in the image interpolation experiment, measured as the MSE between the original input image and the recovered image, by assigning the predicted color for a superpixel to every pixel within it. The x-axis represents the ratio between the number of nodes in $\ccalG_Y$ and $\ccalG_X$. The y-axis shows the MSE defined in \eqref{eq:mse}. Each line corresponds to one of the schemes introduced in Section \ref{SS:imageintr}. The figure represents the mean MSE for 10 realizations of the selection of the pixels in $\ccalV_Y^{tr}$, while the vertical lines represent their standard deviation.}
    \label{fig:img_segm_ny}
\end{figure}

Figure~\ref{fig:img_segm_ny} represents the MSE of the predictions from each architecture as we increase the ratio between the number of superpixels (nodes) in the input and output graphs. To do this, we start with $N_x=N_Y=50$, keep the number of nodes (superpixels) in $\ccalG_X$ fixed, and increase the number of nodes in $\ccalG_Y$ by the ratio shown on the x-axis.
As baselines, we compare our architectures with the ``Mean X'' ($f^\dagger (f_{N_X} (\ccalX))$) and ``Mean Y'' (the target signal and $f^\dagger(f_{N_Y} (\ccalX))$) alternatives.
Note that, since the mean minimizes the MSE, the line labeled as ``Mean Y'' represents a lower bound for the error in this setting.

In the figure, it is noticeable that the only architecture improving its performance and surpassing the dummy solution of using the coarse input as the prediction (``Mean X'') is the IOGCN architecture. Its error decreases as the number of nodes in $\ccalG_Y$ increases, emphasizing the growing importance of leveraging the graph structure (close nodes become more similar, and therefore, the information in the neighbors helps with the prediction). Comparing the alternatives for $\psi^Z_{\bbTheta_Z}$, the architecture that performs best is the learnable linear mapping for the IOGCN architecture (``-W'') and the domain-specific transformation for the IOMLP architecture (``-S'').

\subsection{Fluid flow prediction} \label{SS:fluid}

In this case, we shift our focus to the field of CFD, where the behaviour of a fluid along a pipe or around an object is studied by solving the Navier-Stokes partial differential equations on a mesh.
However, these equations must be solved numerically, and to achieve a sufficient degree of precision, a very fine mesh is required. This results in high computational times, as the finer the mesh, the more costly the simulations become. Data-driven approaches leveraging deep learning have been proposed in the past, showing promising results~\cite{belbute20solvers,bhatnagar2019prediction}.

As in the previous experiment, we create two RAGs, one with a coarse mesh and another with a fine mesh. We then use our architecture to interpolate the data from the coarse mesh (input) to the fine mesh (output) with the same features as the input. More precisely, the features we aim to predict are the fluid speed (in both the x and y axes) and the pressure of the fluid.

\begin{figure}
    \centering
    \includegraphics[width=0.5\textwidth]{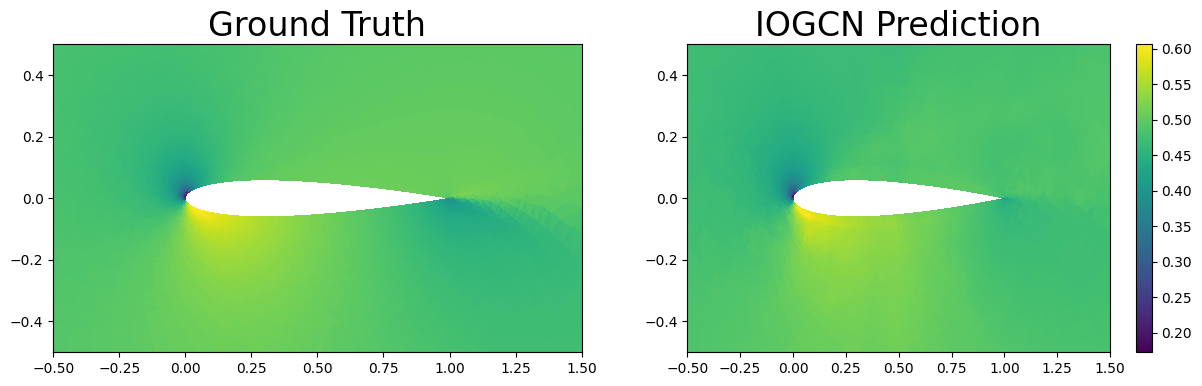}
    \caption{Fluid flow prediction experiment. Two dimensional representation of the ground-truth fluid speed (left) and output generated by the IOGCN scheme (right) for the interpolation experiment and using the linear transformation for $\psi^Z_{\bbTheta_Z}$ (``IOGCN-W'' in Table~\ref{T:cfd_results}).}
    \label{fig:fluid_flow_sample}
\end{figure}

In this case, we consider a fully supervised setting where the training and test sets are created by simulating different configurations of the Mach number and angle of attack for the NACA0012 wing profile, represented in Figure~\ref{fig:fluid_flow_sample} (see~\cite{belbute20solvers} for more details on the simulation setup).
As in~\cite{belbute20solvers}, we carry out two types of experiments. First, we perform an interpolation experiment, where the behaviour (combinations of Mach number and angle of attack) in the training and test datasets is similar. Then, we run a generalization experiment, where the test dataset includes shocks, which occur at high Mach numbers. Shocks were not present in the training set, which was created using low Mach numbers.


Table~\ref{T:cfd_results} lists the prediction performance, measured as the MSE between the predicted and actual signal in the fine mesh. We compare two of our architectures, IOGCN and IOGAT, with several baselines, including the results presented in~\cite{belbute20solvers}, the IOMLP architecture, and a GAT scheme that skips $\psi^X_{\bbTheta_X}$ and implements a GAT over the fine mesh, resulting in the output $\hbY = \psi^Y_{\bbTheta_Y} (\psi^Z_{\bbTheta_Z} (\bbX) | \ccalG_Y)$.
For the alternatives to $\psi^Z_{\bbTheta_Z}$, and similarly to the previous experiment, we implement the learnable linear mapping (``-W'') and the selection domain-specific transformation that selects the embeddings of the nearest nodes (``-S'').

In both the interpolation and generalization experiments, our IOGCN architecture outperforms the alternatives, achieving a lower MSE on the fine mesh.
Specifically, in the interpolation experiment, the learnable mapping in $\psi^Z_{\bbTheta_Z}$ attains the lowest error. Conversely, in the generalization experiment, the domain-specific transformation introduced in the previous section, which assigns values based on proximity in the mesh, performs the best.
This observation may be attributed to the increased complexity of the generalization experiment, where it becomes crucial to leverage as much information as possible, including that embedded in the domain-specific transformation, to achieve optimal results.
The last important observation is that the baselines using GNNs only over the fine mesh, namely GCN and GAT, perform worse than our proposed architecture IOGCN, showcasing the importance of considering the information in both the input and output graphs.

\begin{table}[]
\def\arraystretch{1.2}
\begin{tabular}{c|c|c}
 & Interpolation experiment & Generalization experiment \\ \hline
CFD-GCN~\cite{belbute20solvers}$^*$ & $1.8\cdot 10^{-2}$ & $5.4\cdot10^{-2}$ \\
GCN$^*$ & $1.4\cdot10^{-2}$ & $9.5\cdot10^{-2}$ \\ \hline
IOGCN-W & $\mathbf{6.7\cdot10^{-3}}$ & $4.1\cdot10^{-2}$ \\
IOGCN-S & $1.7\cdot10^{-2}$ & $\mathbf{3.7\cdot10^{-2}}$ \\
IOGAT-W & $8.7\cdot10^{-3}$ & $6.6\cdot10^{-2}$ \\
IOGAT-S & $8.3\cdot10^{-3}$ & $6.2\cdot10^{-2}$ \\ \hline
IOMLP-W & $8.4\cdot10^{-3}$ & $4.0\cdot10^{-2}$ \\
GAT & $1.1\cdot10^{-2}$ & $1.1\cdot10^{-1}$
\end{tabular}
\caption{Fluid flow prediction experiment. MSE associated with the interpolation carried out by each scheme. The best-performing architectures for each experiment are \textbf{boldfaced}. $^*$The MSE values for CFD-GCN and GCN are the ones reported in~\cite{belbute20solvers}.}
\label{T:cfd_results}
\end{table}

\subsection{Self-supervised learning} \label{SS:ccassl} 

In the last experiment, we explore the approach put forth in Section~\ref{S:cca_ssl} and evaluate the proposed framework in the setting of SSL. Specifically, starting from a graph $\ccalG = (\ccalV, \ccalE)$, we consider to have access to two imperfect versions of it, namely: (i) a perturbed version $\ccalG' = (\ccalV, \ccalE')$ with an incomplete edge set, and (ii) an error-free subgraph $\ccalG_s = (\ccalV_s, \ccalE_s)$ with $\ccalV_s \subset \ccalV, \; \ccalE_s \subset \ccalE$. Regarding the signals, for graph $\ccalG'$ we assume to have access to incomplete features, represented as $\bbX' \in \reals^{N \times F}$ where some features are unknown and therefore masked (i.e., columns of $\bbX'$ set to $\bbzero$). For graph $\ccalG_s$, we assume to have access to the signals $\bbY_s = \bbC \bbX \in \reals^{N_s \times F}$, with $N_s=|\ccalV_s|$ and $\bbC \in \{0,1\}^{N_s \times N}$ being a fat sampling matrix that selects the rows of $\bbX$ corresponding to the nodes in $\ccalV_s$. In the experiment, the edges dropped in $\ccalE'$, the features masked in $\bbX'$ and the nodes to keep in $\ccalV_s$ are selected uniformly at random and, to mitigate the effect of these random factors, the simulations are repeated 25 times and the average results are reported. 

Our focus is to learn informative views $\bbZ_s$ of the nodes in $\ccalG_s$ to solve a downstream node classification task, i.e. use $\bbZ_s$ to train a simpler classifier $g_{\bbXi}$ to find the labels $\bbrho \in \reals^{N_s}$ for the nodes in the subgraph $\ccalG_s$.
Previous works dealing with SSL in graphs~\cite{zhang2021from} required the node set $\ccalV$ to remain fixed while modifying the edge set $\ccalE$ and the node features $\bbX$. In the setting introduced, this is equivalent to only using $\ccalG'$, and not being able to leverage the information in $\ccalG_s$, so that one is limited to set $\bbZ_s = \bbC \psi^X_{\bbTheta_X} ( \bbX' | \ccalG' )$. In contrast, when using our framework, we  also leverage the information in $\ccalG_s$ by computing $\bbZ_s = \psi_{\bbTheta_Y}^{Y \; \dagger} (\bbY | \ccalG_s )$, which proves beneficial to the downstream node classification task.

\begin{figure}
    \centering
    \includegraphics[width=0.5\textwidth]{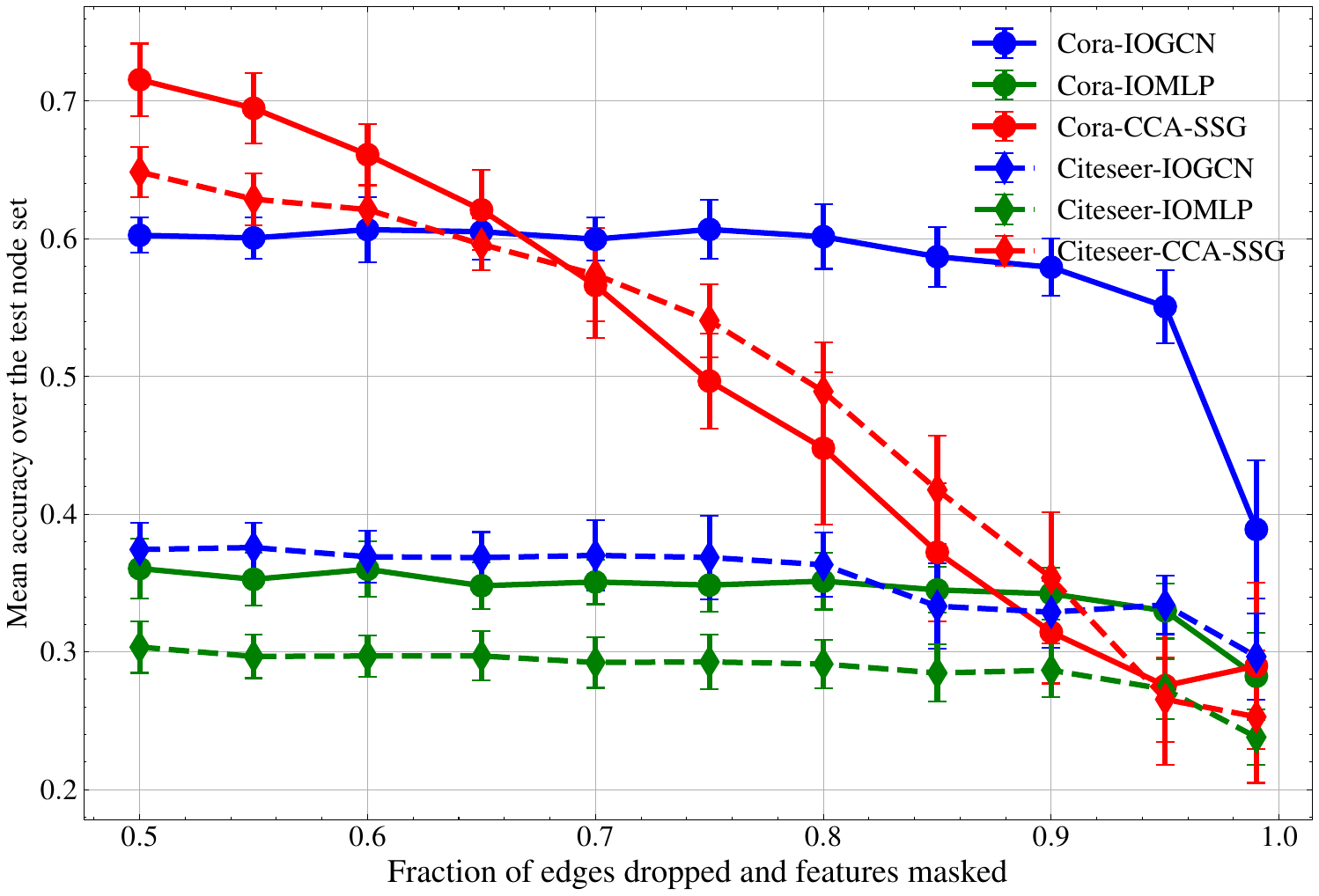}
    \caption{Node classification experiment in the SSL setting described in Section \ref{SS:ccassl}. The Cora and Citeseer datasets are used, and the goal is to predict the labels of a subset of nodes of $\ccalG_s$. Rather than assuming the full graph $\ccalG$ to be known, our setup only has access to two graphs: a perturbed graph $\ccalG'$ where a fraction of the edges are dropped, and an error-free subgraph where only some of the nodes and edges are known. When learning the latent representation of the data, the signals associated with $\ccalG'$ correspond to masked versions of the complete set of features, while the signals associated with $\ccalG_s$ are not masked but do not include labels of nodes not present in $\ccalV_s$. To simplify the experiments, we consider the ratio of dropped edges in $\ccalG'$ and the ratio of masked features in $\bbX'$ to be the same. Different values for this ratio are considered, as indicated on the x-axis. The figure displays the mean accuracy over 25 realizations, while the vertical lines represent their standard deviation.}
    \label{fig:ssl_cca}
\end{figure}

Figure~\ref{fig:ssl_cca} shows the classification accuracy over the test set of nodes for the Cora and Citeseer citation networks, where the nodes in the training, validation, and test sets are selected uniformly at random from $\ccalV_s$. Three schemes are tested: our IOGCN architecture; the IOMLP alternative; and the architecture proposed in~\cite{zhang2021from}, labeled ``CCA-SSG''.
The x-axis shows the perturbation ratio of $\ccalG'$, where the perturbation comes from random edge dropping (we remove edges of the original graph by selecting them uniformly at random) and feature masking (we set the values of the masked features to 0). The value on the x-axis represents the fraction of edges dropped and features masked, which are set to the same value as they both measure the information contained in $\ccalG'$ (the higher the value, the less informative $\ccalG'$ is). As we see, CCA-SSG outperforms our architecture in a setting with minimal perturbations. However, as we increase the perturbations in $\ccalG'$, leveraging the information in $\ccalG_s$ becomes more valuable and IOGCN is able to outperform CCA-SSG, maintaining its performance relatively constant even in a highly perturbed setting.

\section{Conclusion}\label{S:conclusion}

This work presented a deep learning architecture to deal with pairs of signals that are defined over different graphs. While our main focus was on supervised architectures where we are given an input graph signal defined on one graph and the goal is to estimate an output graph signal defined on a different graph, the last part of the paper also discussed the case of deep graph CCA, where the goal is to obtained a reduced dimensionality representation from the two pairs. The architecture proposed to deal with this type of datasets consisted on three main blocks. The first and third blocks were GNNs with no pooling operator. The first GNN was operating over the input graph to process the input signal. Analogously, the GNN in the third block was operating over the output graph to process the output signal. The intermediate block, which operated in a latent space, served a twofold purpose: matching the dimensions on the input and output signals, and incorporating available domain knowledge regarding the relation between the input and output graphs. Various alternatives were proposed for this intermediate block, including low-rank transformations to reduce the degrees of freedom and dimension of the latent space, and non-linear learnable transformations in the form of an MLP. To illustrate the potential value of the architecture, different experimental setups that showcased the utility of leveraging the information in the two graphs were provided. Future work includes generalizing the results to supervised setups with more than two graphs as well as to unsupervised setups where multiple views, each defined over a different graph, are available and the goal is to obtain lower-dimensional representations.


\bibliographystyle{IEEEtran.bst}
\bibliography{biblio}

\end{document}